\documentclass[acmsmall]{acmart} 
\usepackage{multirow}
\usepackage{algorithm}
\usepackage{algorithmic}

\AtBeginDocument{%
  \providecommand\BibTeX{{%
    \normalfont B\kern-0.5em{\scshape i\kern-0.25em b}\kern-0.8em\TeX}}}

\acmDOI{XXXXXXX.XXXXXXX}

\begin{document}

\title{An Imitative Reinforcement Learning Framework for Pursuit-Lock-Launch Missions}

\author{Siyuan Li}
\authornote{Corresponding author}
\email{siyuanli@hit.edu.cn}
\affiliation{%
  \institution{Harbin Institute of Technology}
  \country{China}
  \postcode{150001}
}

\author{Rongchang Zuo}
\email{2021110788@stu.hit.edu.cn}
\affiliation{%
  \institution{Harbin Institute of Technology}
  \country{China}
  \postcode{150001}
}

\author{Bofei Liu}
\email{bofeiliu029@gmail.com}
\affiliation{%
  \institution{Harbin Institute of Technology}
  \country{China}
  \postcode{150001}
}

\author{Yaoyu He}
\email{heyy20@mails.tsinghua.edu.cn}
\affiliation{
\institution{Tsinghua University}
\country{China}
\postcode{100084}
}

\author{Peng Liu}
\email{pengliu@hit.edu.cn}
\affiliation{%
  \institution{Harbin Institute of Technology}
  \country{China}
  \postcode{150001}
}

\author{Yingnan Zhao}
\email{zhaoyingnan@hrbeu.edu.cn}
\affiliation{%
  \institution{Harbin Engineering University}
  \country{China}
  \postcode{150006}
}


\renewcommand{\shortauthors}{Siyuan Li, et al.}

\begin{abstract}
Unmanned Combat Aerial Vehicle (UCAV) {Within-Visual-Range (WVR) engagement}, referring to a fight between two or more UCAVs at close quarters, plays a decisive role on the aerial battlefields. 
With the development of artificial intelligence, {WVR engagement} progressively advances towards intelligent and autonomous modes.
However, autonomous {WVR engagement} policy learning is hindered by challenges such as weak exploration capabilities, low learning efficiency, and unrealistic simulated environments.
To overcome these challenges, we propose a novel imitative reinforcement learning framework, which efficiently leverages expert data while enabling autonomous exploration. The proposed framework not only enhances learning efficiency through expert imitation, but also ensures adaptability to dynamic environments via autonomous exploration with reinforcement learning.
Therefore, the proposed framework can learn a successful policy of `pursuit-lock-launch' for UCAVs.
To support data-driven learning, we establish an environment based on the Harfang3D sandbox.
The extensive experiment results indicate that the proposed framework excels in this multistage task, and significantly outperforms state-of-the-art reinforcement learning and imitation learning methods. 
Thanks to the ability of imitating experts and autonomous exploration, our framework can quickly learn the critical knowledge in complex aerial combat tasks, achieving up to a $100\%$ success rate and demonstrating excellent robustness.
\end{abstract}





\maketitle

\section{Introduction}

In recent years, emerging artificial intelligence technologies, such as deep learning have made remarkable advancements in the military field. The increasing significance of artificial intelligence applications in military operations has propelled modern warfare toward intelligent development \cite{van2019framework,benaskeur2010corals}. 
Unmanned combat aerial vehicles (UCAVs), as key elements in aerial combat, have become essential in modern military and defense, demonstrating outstanding combat performance in tasks such as aerial strikes, reconnaissance, electronic warfare, and target tracking. Autonomous control of UCAVs has gradually become a primary research focus \cite{hu2022autonomous,chamola2021comprehensive,bell2017evaluating}, where {WVR engagement} is one of the most important tasks for UCAVs.
{WVR engagements} refer to fights between two or more UCAVs usually at close quarters, which typically involve three subtasks: pursuing, locking, and launching. This paper mainly focuses on autonomous control policy learning for this pursuit-lock-launch task between two UCAVs.

Recently, more researchers have investigated reinforcement learning (RL) \cite{sutton2018reinforcement} techniques to achieve autonomous policy learning for UCAVs \cite{kim2003autonomous,wang2022research,fu2022uav,li2023manoeuvre,fan2022air,cao2023autonomous}. Different from supervised learning methods, RL enables the control agent for the UCAV to gradually learn policies through online interactions with the environment. 
Considering the dynamic property of the {WVR engagement} environment, RL is suitable for such a learning problem \cite{piao2024discovering}, as it supports online exploration and therefore can better adapt to dynamic changes in the environments.
However, {WVR engagement} policy learning is quite a complex problem, which involves multiple control dimensions, e.g., the direction of flight, target lock, and missile launch. 
Due to the trial-and-error nature of RL agents, \textbf{the sample efficiency of RL algorithms in such complex learning scenarios is low}, leading to poor success rates of {WVR engagements}. 

Besides the learning frameworks, the simulated learning environment is substantially important as well.
Since directly conducting {WVR engagement} policy learning in real environments is impossible, an effective and high-fidelity simulator is urgently required.
However, \textbf{the simulated environments in previous works suffer from the oversimplified action spaces and task setting.}
For example, the action space in \cite{fan2022air} is simplified into a discrete predefined maneuver library;  \cite{fu2022uav} confines the action space to a two-dimensional simulation environment; \cite{wang2022research, li2023manoeuvre, cao2023autonomous} merely use part of the parameters of the UCAV flight dynamics equations as the action space.
As for the task setting, \cite{kim2003autonomous} solely focuses on the flight control of UCAV without launching missiles; \cite{wu2023intelligent} tackles missile evasion of UCAV without flight control; \cite{hsu2018distributed, choi2019deep, zhu2017target, kahn2018self} aim to achieve the objectives of route planning; \cite{haarnoja2018soft, fu2022uav} aim to accomplish the pursuit subtask; \cite{li2023manoeuvre} completes the locking of the opponent based on rule-based pursuing.
The flight dynamics model and task setting in JSB-sim \cite{berndt2004jsbsim} are less realistic than ours, since this simulator does not integrate a missile system, burdening researchers to manually simulate missiles or utilize existing open-source repositories.
To successfully shoot down the opponent in a pursuit-lock-launch task, UCAV needs to accomplish three subtasks at least: pursuiting, locking, and launching.
Specifically, these include swiftly pursuing the opponent through the optimal path, locking onto the target within a proper range, and launching missiles once the opponent is locked.

To address the aforementioned challenges in a pursuit-lock-launch policy learning, we introduce a novel imitative reinforcement learning framework that integrates imitation learning with autonomous exploration by the RL agent. By imitating expert data, learning efficiency is improved, while autonomous exploration by the RL agent facilitates adaptation to dynamic environments. 
In this work, the UCAV pursuit-lock-launch problem is formulated as a Markov Decision Process (MDP). 
To alleviate the limitations of previous dogfight simulators,  we construct a realistic UCAV simulation environment based on the Harfang3D sandbox \cite{ozbek2022harfang3d}.
The constructed learning environment features a high-fidelity action space design that enables control over UCAV directional rudders, elevators, and ailerons, making it more realistic than prior works that employed simplified dynamics equation parameters for control.
Beyond the action space, the dynamic model in the constructed simulator is also of high fidelity, which is built according to the flight dynamics models of real aircraft.
As a replication of the {WVR engagement} process, the simulated pursuit-lock-launch task consists of three stages, i.e., pursuit, lock, and launch, providing a more complex replication of real {WVR engagement} processes than previous single-stage works.
As an evaluation, we conduct experiments in the constructed {WVR engagement} environment.
The experiment results demonstrate that the training environment constructed in this paper has high fidelity, effectively simulating the UCAV {WVR engagement} process and better aligning the physical model with real-world scenarios. The proposed learning framework performs excellently in the UCAV multi-dimensional decision-making task, significantly outperforming state-of-the-art deep RL and imitation learning algorithms, achieving higher learning efficiency, more effective air combat policies, and higher success rates. The contributions of this work can be summarized as follows.
\begin{itemize}
    \item \textbf{Problem Formulation:} We construct a simulated {WVR engagement} environment with high fidelity, which contains multiple stages: pursuit, lock, and launch.  
    \item \textbf{Learning Algorithm:} We propose a novel imitative reinforcement learning framework for UCAV {WVR engagement} that enhances learning efficiency and adapts to complex environments.
    \item \textbf{Empirical Improvement:} The proposed framework significantly surpasses the state-of-the-art RL and imitation learning methods in the pursuit-lock-launch problem.
\end{itemize}

This work is organized as follows. In Section 2, we introduce preliminary knowledge, and then in Section 3, we review the related works. After that, in Section 4, we formulate the pursuit-lock-launch problem as an MDP. In Section 5, we present the proposed framework for pursuit-lock-launch policy learning. Next, in Section 6, we conduct extensive experiments, and the results demonstrate that the proposed framework substantially outperforms the baselines. Finally, in  Section 7, we conclude and point out the possible future directions.

\section{Preliminaries}

In this section, we review the preliminary knowledge about the Harfang3D Sandbox simulator, the Markov decision process, and the actor-critic framework.

\subsection{Harfang3D Sandbox}

\begin{figure}[htbp]
    \centering
    \includegraphics[width=\linewidth]{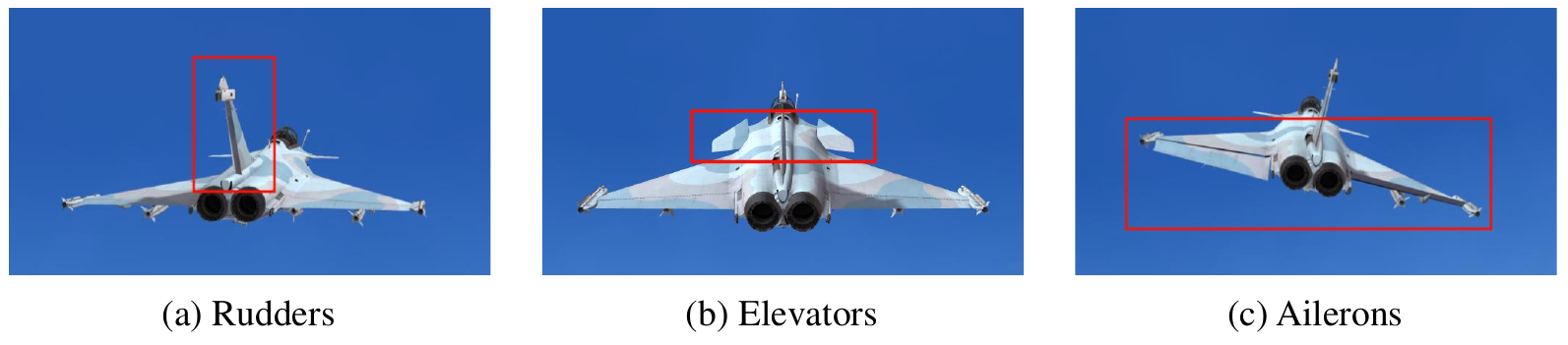}
    \caption{Diagram of aircraft attitude control.}
    \label{fig1}
\end{figure}

We establish a simulation platform based on the Harfang3D sandbox \cite{ozbek2022harfang3d}, which is a customizable and flexible tool for combat aircraft dogfights. This platform facilitates the manipulation and access to flight dynamics models and environment states. The sandbox provides simulated aircraft of five different models, each equipped with unique missile configurations, and a limited missile supply. Therefore, the aircraft must lock onto a target before launching a missile to avoid missing the target and consequently losing a missile, which increases the task difficulty. 
Within this simulation environment, information about environment states can be obtained, including aircraft position, attitude, heading angle, and health as well as target lock status, thrust, lock angle, and missile loading.
In particular, the sandbox also offers various unmanned aircraft control methods for components including rudders, elevators, ailerons, missile launch, engine thrust, landing gear, and flaps. As illustrated in Figure \ref{fig1}, rudders control the aircraft heading, elevators control the aircraft pitch, and ailerons control the aircraft roll. These three controls enable basic transformations of the aircraft's attitude.

\subsection{Markov Decision Process}

The Markov Decision Process (MDP) \cite{garcia2013markov} is the theoretical foundation of RL, which provides a formal description of the learning environment of the agent.
It is noteworthy that, since the intelligent agent cannot access complete details of the opponent's information during aerial combat, modeling the scenario as a Partially Observable Markov Decision Process (POMDP) appears more theoretically justified. However, we model the UCAV {WVR engagement} environment as an MDP to simplify the problem in this paper. The MDP is represented as a tuple $(S, A, P, R, \gamma)$, where $S$ is the state space of the environment, $A$ is the action space of the agent, $P$ is the state transition probability, with $P(s_{t+1} | s_t, a_t)$ indicating the probability of transitioning from state $s_t$ to state $s_{t+1}$ when taking action $a_t$ at time $t$, $R$ is the reward function, representing the immediate reward obtained by the intelligent agent after taking action $a$, and $\gamma$ is the reward discount factor, with $\gamma \in [0, 1)$  and a $\gamma$ value closer to 1 indicating that the intelligent agent focuses more on long-term rewards. Reinforcement learning aims to learn policy function $\pi: S \rightarrow A$  that maximizes discounted reward  $\mathbb{E}_{P, \pi}[\sum_{t=1}^T\gamma^tR(s_t, a_t)]$, where $T$ is the total timestep in an episode.

\subsection{Actor-Critic Framework}

The actor–critic framework \cite{konda1999actor} is a policy optimization paradigm in RL that simultaneously optimizes both the value and policy functions, exhibiting good learning performance in sequential decision problems with continuous action spaces.
The actor represents the policy $\pi$  of the agent, which specifies the actions to take at each state. It aims to learn an optimal policy that maximizes the expected cumulative reward. During training, the actor adjusts its policy based on feedback from the critic, gradually improving the policy performance. The critic estimates the value of each action to assess the quality of the policy and assists the actor in refining its policy.
The actor–critic framework is based on policy gradient methods, combining back-propagation with value function methods to enhance the policy performance. Based on the actor-critic framework, the agent can learn a policy and corresponding value function during interaction with the environment, enabling high-level decision-making and intelligent behavior.

\section{Related Work}

With the rapid development of deep learning, the control for UCAVs has become more autonomous.
Deep learning enables UCAVs to both execute fixed programs and exhibit learning capabilities. In recent years, deep learning techniques such as supervised learning, imitation learning, and RL have been widely used to address tactical decision-making problems in UCAV dogfights.
In this section, we discuss the related works from four views: supervised learning methods, imitation learning methods, reinforcement learning methods, and methods of combining RL and imitation learning.

\subsection{Supervised Learning}
Supervised learning is predominantly used in UCAV dogfights to predict the states after executing actions, providing a basis for action selection. Zhang and Huang \cite{zhang2020maneuver} proposed a deep learning method integrating state evaluation and action selection for making flight action decisions. The state evaluation function scores the next state after executing an action and selects the action that generates the highest score. This approach enables autonomous flight control of UCAVs in one-on-one aerial combat. Li et al. \cite{li2022decision} trained a long short-term memory network with time-series data, mapping historical state sequences onto flight control parameters. They used a stacked sparse autoencoder to reduce data redundancy and accelerate decision-making.

\subsection{Imitation Learning}
Imitation learning \cite{hussein2017imitation} 
is a machine learning paradigm where an agent tries to learn a policy by mimicking expert behavior.
Imitation learning methods include 1) behavior cloning (BC), 2) inverse reinforcement learning, and 3) generative adversarial imitation learning.

(1)	BC \cite{ly2020learning} involves training a model on expert demonstration data to resemble the expert behaviors. Sandström et al. \cite{sandstrom2022fighter} adopted BC using control signals from human pilots to train a neural network, enabling the model to mimic the behaviors of human pilots for aircraft flight.

(2)	Inverse reinforcement learning \cite{arora2021survey} estimates the reward function from expert demonstrations and uses it to guide the training of intelligent agents in reinforcement learning. By extracting the hidden reward function from expert data, the system can understand task objectives from expert demonstrations. Kong et al. \cite{kong2020uav} employ a reward shaping method to address the problem of sparse rewards in reinforcement learning and use maximum entropy inverse reinforcement learning to obtain a shaped reward function, thereby accelerating convergence in training.

(3)	Generative adversarial imitation learning \cite{ho2016generative} uses adversarial generative networks to learn hidden policies from expert demonstrations, imitating expert actions. Wang and Wei \cite{wang2022research} applied this technique to address the problem of sparse rewards in reinforcement learning. They designed internal rewards to stimulate intelligent agents to explore autonomously, thus expediting training.

\subsection{Reinforcement Learning}
Reinforcement learning introduces more autonomous and adaptive methods for dogfight \cite{wang2022research,fu2022uav,li2023manoeuvre,fan2022air,cao2023autonomous}. The RL agent actively interacts with the environment and learns optimal policies through trial and error. The RL agent adjusts behavior based on received reward signals, gradually improving performance to adapt to the constantly changing environment. 
Fan et al. \cite{fan2022air} propose an autonomous maneuver decision method for UCAV based on the Asynchronous Advantage Actor-Critic (A3C) \cite{mnih2016asynchronous} algorithm. Relying on a library of seven predefined actions, they train UCAV to gain an advantageous position in aerial combat, and the training efficiency is improved through a multi-threaded asynchronous mechanism. 
Cao et al. \cite{cao2023autonomous} integrate RL with game theory, using Double Deep Q-Network (DDQN) \cite{van2016deep} algorithm to train a three-degree-of-freedom UCAV to execute seven basic aerial combat maneuvers and maintain a superior position in air combat. 
Wang et al. \cite{wang2022research} address the issue of multi-UCAV aerial combat by designing reward functions that incorporate prior knowledge. They utilize the Multi-Agent Deep Deterministic Policy Gradient (MADDPG) \cite{lowe2017multi} algorithm to overcome the challenges associated with high-dimensional, continuous action spaces that traditional reinforcement learning methods face. This approach enable multiple UCAVs to maintain advantageous positions in aerial combat. 
Fu et al. \cite{fu2022uav} use Deep Deterministic Policy Gradient (DDPG) \cite{lillicrap2015continuous} algorithm to train a two-degree-of-freedom UCAV to complete tracking and evasion tasks in UCAV dogfight, introducing imitation learning to address the inefficiency of exploring different scenarios during the training phase of reinforcement learning algorithms. 
Li et al. \cite{li2023manoeuvre} build upon the Soft Actor-Critic (SAC) \cite{haarnoja2018soft} method, utilizing expert knowledge to prevent SAC from getting stuck in local optima and accelerating the training process. This work trains a three-degree-of-freedom UCAV to perform pursuit and lock decisions in UCAV dogfight, but has not addressed the weapon usage problem.
Zhang et al. \cite{zhang2023multi} propose to expand dogfight decision-making from a single dimension to multiple dimensions, including target pursuit, target lock, and weapon use. This work employs the SAC algorithm for policy learning and establishes a set of meta-policies. However, this work is limited to a two-dimensional environment, and the weapon meta-policies are manually defined.

\subsection{Combining Reinforcement Learning and Imitation Learning}
To the best of our knowledge, this work is the first to investigate UCAV control policy learning via combining RL and imitation learning. In other domains, similar techniques have been tried. For example, in the offline RL domain, Fujimoto et al. \cite{fujimoto2021minimalist} propose to combine the TD3 \cite{fujimoto2018addressing} method and the BC method. In contrast to this offline learning method, the proposed approach can leverage online exploration to further enhance policy learning. 
In the quantitative trading domain, previous researchers combine RL with imitation learning to solve the challenging portfolio management, market making, and order execution problems \cite{imm, macmic, niu2022metatrader}.

Although deep learning methods, especially deep RL, have made significant progress in UCAV dogfights, current research still faces several challenges, such as unrealistic simulation environments, oversimplified task design, low algorithmic sample efficiency, and inadequate exploration capabilities. 
To address these issues, we construct a realistic UCAV {WVR engagement} simulation environment based on the Harfang3D sandbox, which utilizes a UCAV control mechanism aligned with real operational procedures, and design a comprehensive pursuit-lock-launch task that incorporates multiple decision dimensions. 
Furthermore, we propose a novel imitative reinforcement learning framework that enhances learning efficiency through expert imitation.
Beyond that, the proposed approach can
improve the policy's adaptability to dynamic environments through autonomous exploration, and therefore mastering complex multi-dimensional decision policies in UCAV pursuit-lock-launch tasks.

\section{Problem Formulation}

\begin{figure}[htbp]
    \centering
    \includegraphics[width=0.8\linewidth]{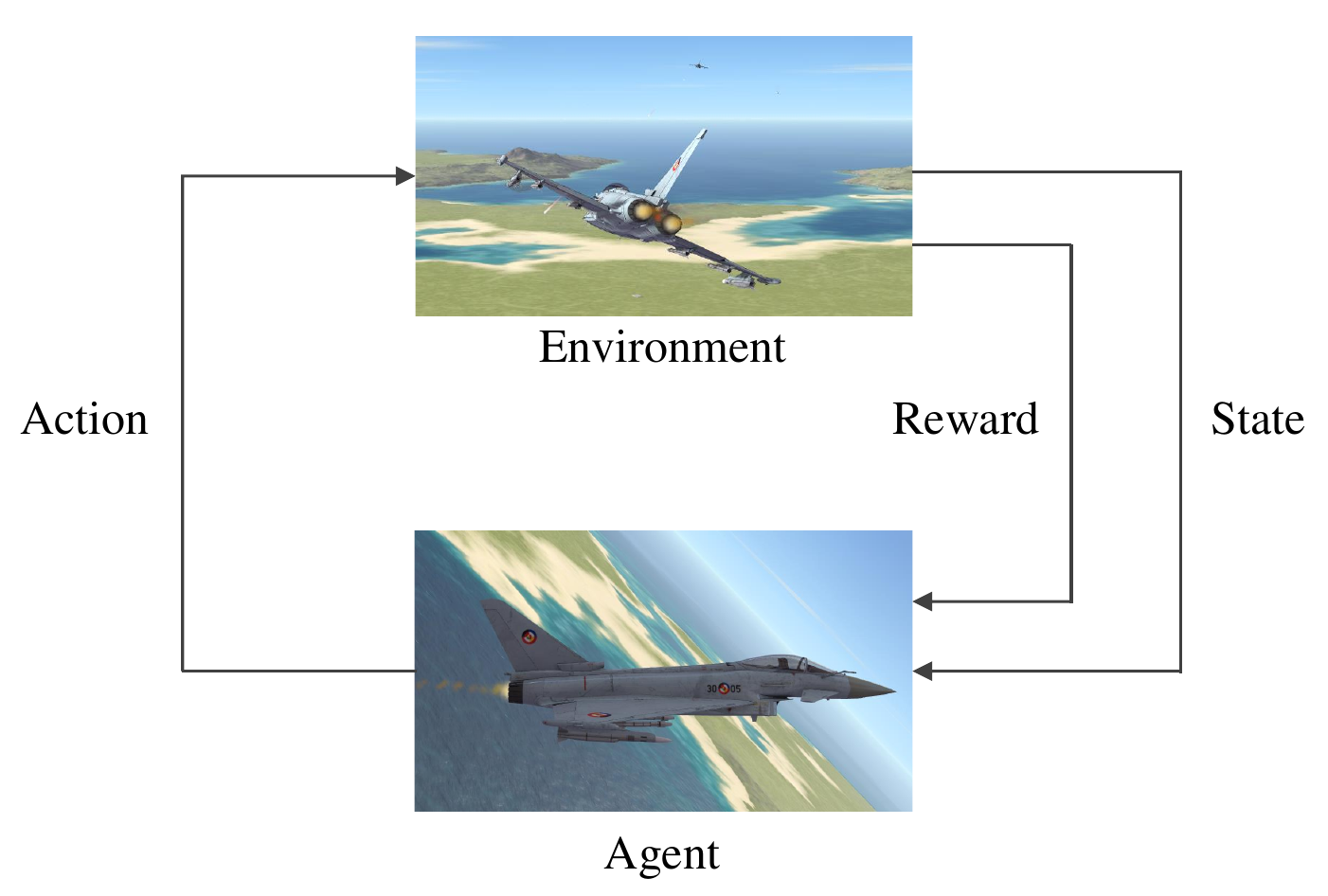}
    \caption{The formulated MDP environment.}
    \label{fig2}
\end{figure}

We establish a realistic simulation environment for UCAV {WVR engagement} policy learning based on the Harfang3D sandbox \cite{ozbek2022harfang3d} and model the `pursuit-lock-launch' tasks in a {WVR engagement} as an MDP, which is shown in Figure \ref{fig2}.

\subsection{Task Description}

\begin{figure}[htbp]
    \centering
    \includegraphics[width=0.8\linewidth]{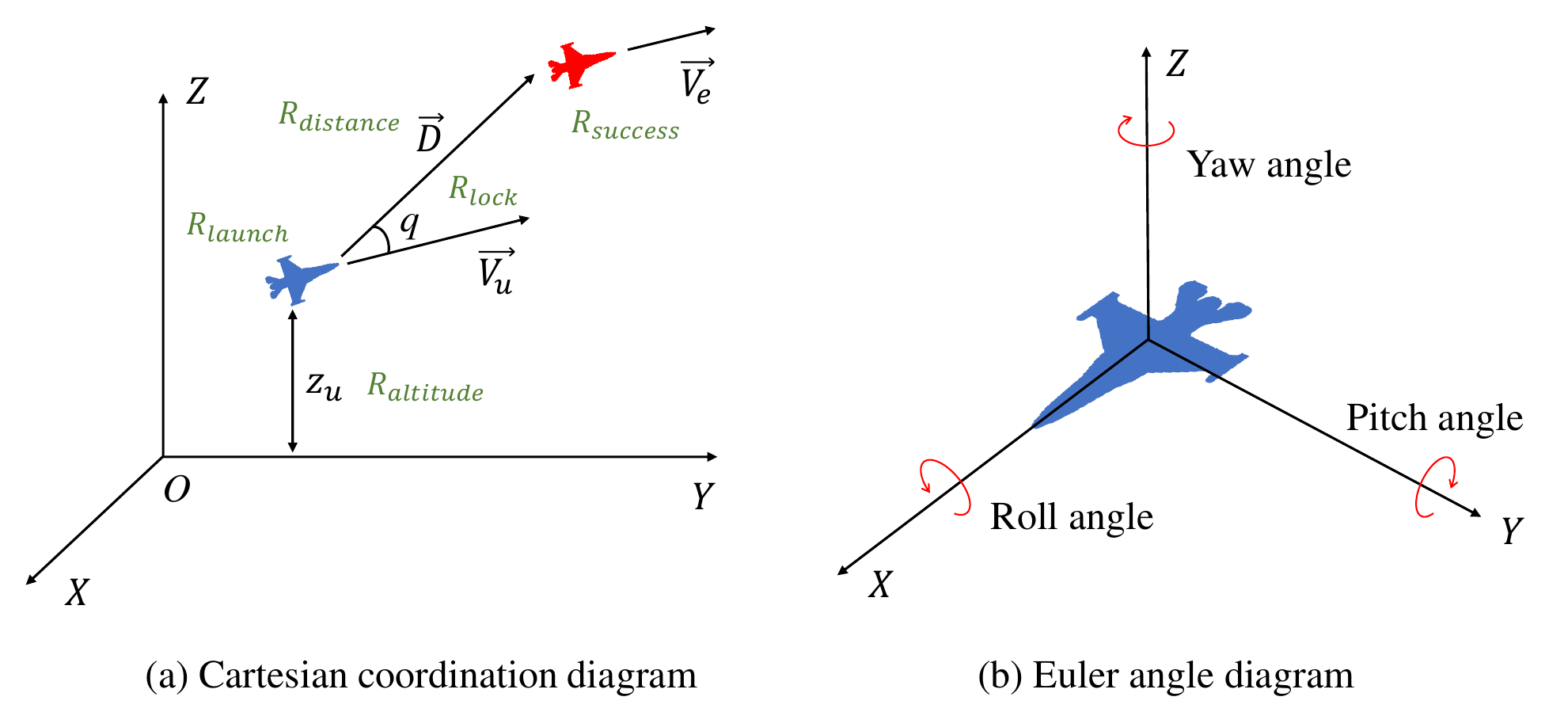}
    \caption{An illustration of {WVR engagement} rewards. The blue aircraft denotes our aircraft, and the red one denotes the opponent aircraft.}
    \label{fig5}
\end{figure}

In this subsection, we describe the UCAV {WVR engagement} task, which is divided into three consecutive stages: target pursuit, target lock, and missile launch.
The initial position of the opponent aircraft is fixed, while the initial position of our aircraft is random, with a relative distance greater than the lock range between them.
Both our aircraft and the opponent aircraft are initialized in the air.
The opponent aircraft have three modes: flying forward with a constant speed, serpentine maneuvering, and circling maneuvering.
In contrast, our aircraft is controlled intelligently through a neural network. If our aircraft catches up with the opponent aircraft, locks onto it, launches a missile, and finally shoots down the opponent aircraft, the mission is successful. To ensure that the agent can successfully and accurately complete this task, it needs to master at least three policies of three different stages. Successful target lock requires meeting lock conditions and maintaining them for $5$ seconds, as given below.
\begin{equation}
    100 < |\vec{D}| < 3000\text{ and } q< 15^\circ,
\end{equation}
where $|\vec{D}|$ represents the relative distance between our aircraft and the opponent aircraft, and $q$ is the angle shown in Figure \ref{fig5}(a). 

\subsection{State Space}

As shown in Figure \ref{fig5}, within the same Cartesian coordinate system, the coordinates of our aircraft are represented as $(x_u,y_u,z_u)$, and those of the opponent aircraft are denoted as $(x_e,y_e,z_e)$. The relative position vector is denoted as $\vec{D}$.$\vec{V_u}$ and $\vec{V_e}$ are the velocity vectors of our aircraft and the opponent aircraft, respectively. The angle $q$ between the velocity vector of our aircraft and the relative position vector is the locking angle, which is calculated as
\begin{equation}
    q = \arccos \left( \frac{\vec{D} \cdot \vec{V_u}}{|\vec{D}| |\vec{V_u}|} \right)\times \frac{180}{\pi}
\end{equation}

The state space is comprised of 13 variables: the Cartesian coordinate differences between our aircraft and the opponent aircraft $(\Delta x, \Delta y, \Delta z)$, the Euler angle of our aircraft, locking angle, locking status, missile status, the Euler angle of opponent aircraft, and opponent aircraft health.

\subsection{Action Space}

The action space consists of four types of actions: rudder, elevator, and aileron control as well as missile launch control variable. The aircraft carried only one missile, thus having one opportunity to shoot down the opponent aircraft. The rudder, elevator, and aileron control are continuous variables, and we normalize them to a range of $(-1, 1)$. The missile launch control variable is discrete, with $-1$ representing no launch and $1$ representing a launch. This hybrid action space, consisting of both continuous and discrete variables, presents increased challenges in pursuit-lock-launch policy learning, but resembles the real-world scenarios more.

\subsection{Reward Function}

The reward function is a critical component for applying reinforcement learning to real-world problems. In the UCAV pursuit-lock-launch task, the operated aircraft pursues the opponent aircraft, locks its aim, and launches a missile within an extremely short time. To accomplish this task, we propose the following reward function with five terms:
\begin{equation}
    R = R_{distance} + R_{lock} + R_{success} + R_{altitude} + R_{launch}.	
\end{equation}
The five kinds of rewards are labeled in Figure \ref{fig5} near the corresponding determining variables of these rewards.

$R_{distance}$ is a reward function designed to ensure that the operated aircraft can catch up with the opponent aircraft. This reward calculates the distance between the operated and opponent aircraft and uses a negative distance value as the reward function defined as: 
\begin{equation}
    R_{distance} = -10^{-4}|\vec{D}|.
    \label{eq2}
\end{equation}
The reward in Eq.\eqref{eq2} encourages the operated aircraft to swiftly close the distance with the opponent aircraft for subsequent target lock.
 
$R_{lock}$ is the reward that incentivizes the operated aircraft to lock onto the opponent aircraft. It uses the negative locking angle value, where a smaller locking angle provides a higher reward. The reward function is given by:
\begin{equation}
    R_{lock} = -10q,
\end{equation}
where $q$ is the locking angle between the velocity vector of our aircraft $\vec{V}_u$ and the relative position vector $\vec{D}$. 

$R_{success}$ is the final reward for achieving success in air combat. This reward is given only when the missile successfully hits the opponent, as follows:
\begin{equation}
    R_{success} = \left\{
    \begin{array}{lr}
        800, & \text{if $B_s =1$} \\
        0, & otherwise 
    \end{array}
    \right.
\end{equation}
where $B_s$ is a Boolean value set to $1$ (true) when the missile launched by our aircraft hits the opponent. $R_{success}$ promotes the aircraft to execute the correct launch action, only providing the final reward only when the opponent is shot down by a missile, with no reward given if the opponent is shot down by ramming.

$R_{altitude}$ constrains the operated aircraft exploration. Penalties are imposed when the aircraft flies at excessively high or low altitudes as follows:
\begin{equation}
    R_{altitude} = \left\{
    \begin{array}{lr}
        -4, & \text{if $z_u >7000$ or $z_u < 2000$} \\
        0, & otherwise
    \end{array}
    \right.
\end{equation}
where $z_u$ is the altitude of our aircraft as shown in Figure \ref{fig5}. 

$R_{launch}$ is the penalty for missile launches. Given that only one missile is available, this penalty encourages the operated aircraft to minimize unnecessary launches. The function is given by
\begin{equation}
    R_{launch} = \left\{
    \begin{array}{lr}
        -6, & \text{if $B_f =1$} \\
        0, & otherwise
    \end{array}
    \right.
\end{equation}
where $B_f$ is a Boolean value set to $1$ (true) when our aircraft launches a missile.

Among the five reward terms,
$R_{distance}$, $R_{lock}$, and $R_{success}$ facilitate the smooth completion of pursuit, lock, and launch, respectively, while $R_{altitude}$ and $R_{launch}$ constrains the exploration space of the operated aircraft to achieve effective exploration.

\section{Methodology}

\begin{figure}[htbp]
    \centering
    \includegraphics[width=0.8\linewidth]{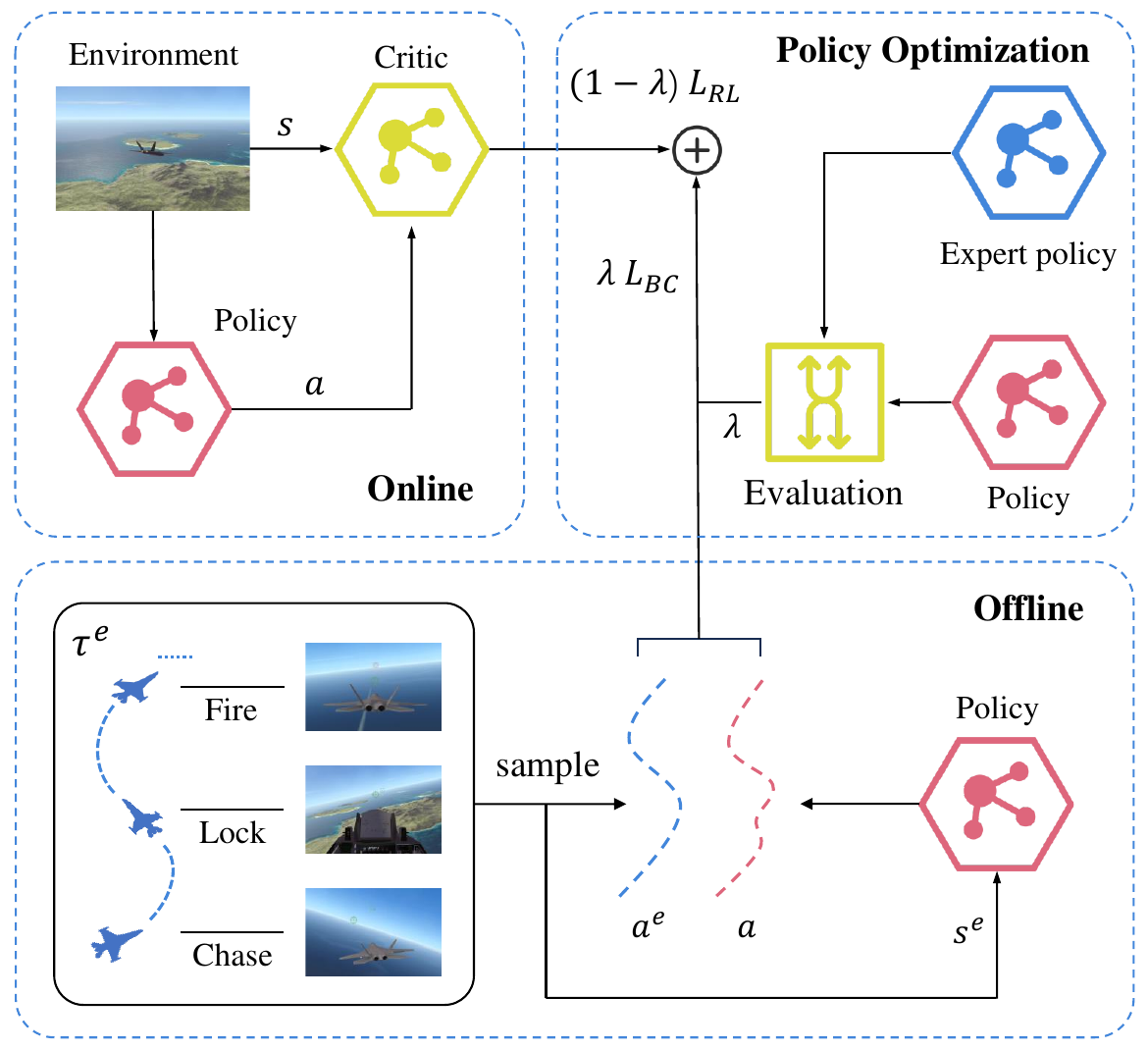}
    \caption{The proposed learning framework for {WVR engagements}.}
    \label{fig7}
\end{figure}

To accomplish the multistage and complex control tasks of pursuing an opponent aircraft, locking onto the opponent, launching a missile, and ultimately shooting down the opponent, we propose an imitative reinforcement learning method that integrates expert knowledge learning with the autonomous exploration of reinforcement learning. As illustrated in Figure \ref{fig7}, the proposed method employs the actor–critic network for learning while leveraging expert data to constrain learning, achieving efficient policy learning. The algorithm details are outlined below.

The actor–critic framework serves as the foundation for pursuit-lock-launch policy learning. 
{Inspired by the TD3 algorithm \cite{fujimoto2018addressing}, to mitigate the overestimation of the critic, we adopt a double Q-network architecture ($Q_{\phi_1}$ and $Q_{\phi_2}$) \cite{van2016deep}.}
Inspired by the RL objective shown in Eq.\eqref{eq6}, the actor network $\theta_{actor}$ is trained to maximize the smaller Q value between the two Q values. 
\begin{equation}
    L_{RL} = -E_{s\sim B}[min_{i=1, 2}Q_{\phi_i}(s, \pi(s; \theta_{actor}))],
    \label{eq6}
\end{equation}
where $L_{RL}$ denotes the RL loss for the actor network, $B$  represents the replay buffer, $\pi(s;\theta_{actor})$  denotes the policy network with parameter $\theta_{actor}$, and $Q(s,a)$  represents the estimated Q value by the critic network. Minimizing loss $L_{RL}$  to update the parameters of the actor network can achieve pursuit-lock-launch policy learning.
{The detailed procedure of neural network optimization includes random sampling a minibatch from the replay buffer $B$, and then optimizing $\theta_{actor}$ with stochastic gradient descent. This is standard in the deep learning domain, and it can be regarded as optimizing the expectation in Equation \eqref{eq6}. For the rest loss functions, the same procedure is adopted.}

However, relying solely on the RL objective likely leads to low sample utilization, as the RL agent explores by trial and error in the environment, thus requiring substantial samples to discover successful policies. In contrast, expert data contain rich prior knowledge. Hence, we introduce a BC term,  $L_{BC}$, as shown in Eq.\eqref{eq7} for the agents to imitate expert policies and efficiently learn key knowledge. BC loss $L_{BC}$ is defined as 
\begin{equation}
    L_{BC} = E_{(s^e,a^e)\sim \mathcal{T} ^e}||a^e - \pi(s^e;\theta_{actor})||^2,
    \label{eq7}
\end{equation}
where $\mathcal{T}^e$  denotes the expert trajectory obtained through artificial intelligence (AI) mode in Harfang3D sandbox and $(s^e,a^e)$  denotes the expert state–action pairs. 
{The proposed approach assumes that an effective expert dataset $\mathcal{T}^e$  is provided. Without expert data, the proposed approach is downgraded to conventional RL methods.}

The proposed learning objective, as expressed in Eq.\eqref{eq8}, supports agents in autonomous exploration and improves their performance for critical actions in tasks (e.g., launch) to approach expert-level capabilities.
\begin{equation}
    \theta_{actor}\leftarrow argmin_{\theta_{actor}}[(1-\lambda)L_{RL}+\alpha\lambda L_{BC}],
    \label{eq8}
\end{equation}
{where  $\alpha$  is a fixed weight to balance the magnitude $L_{RL}$ and $L_{BC}$,
which is key to learning performance
, and $\lambda$ is a weight to balance the two loss terms.} {We employ the most straightforward approach to set $\alpha$ : empirically determining the approximate magnitudes of both losses through several preliminary training episodes, then calculating $\alpha$ to equilibrate their scales.}
In this work, we propose to use the loss function $(1-\lambda)L_{RL}+\alpha\lambda L_{BC}$ to update the actor network, so that we can take advantage of both RL and imitation learning. However, balancing these two learning objectives poses a significant challenge. To address this issue, we develop two mechanisms to set $\lambda$: linear $\lambda$ and adaptive $\lambda$.

\textbf{Linear $\lambda$:} The initial value  for $\lambda$ is $0.5$, and it gradually decreases with the training round $e$ as follows:
\begin{equation}
    \lambda = 0.5-e\times k,
\end{equation}
where $k$ is a constant, and we set $k$ as $0.0002$ in the experiments.

\textbf{Adaptive $\lambda$:} The weight $\lambda$ is automatically updated at each round by comparing the value functions of the current policy and expert policy as follows:
\begin{equation}
    \lambda = \left\{
    \begin{array}{lr}
        1, &\text{if $Q_{\phi_1}(s, \pi^e(s)) >  Q_{\phi_1}(s, \pi(s; \theta_{actor}))$ or $Q_{\phi_2}(s, \pi^e(s)) >  Q_{\phi_2}(s, \pi(s; \theta_{actor}))$}  \\
        0, &otherwise
    \end{array}
    \right.
    \label{eq12}
\end{equation}
where $\pi^e(s)$ is the expert policy network pre-trained with the BC method. 
In the actor-critic framework, the actor and critic form a training unit, with the critic evaluating the actor's actions to guide its updates. In the proposed learning framework, the critic evaluates both the learned policy $\pi(s;\theta_{actor})$ and the expert policy $\pi_{BC}$ as well. When the Q-value of the learned policy surpasses that of the expert policy, it indicates that the learned policy has absorbed or even surpassed expert knowledge. At this point, reducing reliance on the expert policy and instead emphasizing the exploration and optimization of $\pi(s;\theta_{actor})$ can encourage the agent to explore novel actions, aiding in optimizing long-term performance and freeing the policy from being constrained by the static expert knowledge.
In Eq.\eqref{eq12}, the agent imitates the expert when the Q value of the expert action exceeds that of the learned policy, reflecting optimism toward expert data.

\begin{algorithm}
    \caption{Imitative Reinforcement Learning Framework for Pursuit-Lock-Launch Task}
    \label{alg3} 
    \begin{algorithmic}[1]
        \STATE Initialize the actor network with $\theta_{actor}$ and the critic networks with $\phi_1$ and $\phi_2$ 
        \STATE Initialize the target networks $\phi_1'\leftarrow \phi_1$, $\phi_2'\leftarrow \phi_2$, $\theta_{actor}' \leftarrow \theta_{actor}$
        \STATE Initialize weight $\lambda$, replay buffer $B$, and expert trajectories $\mathcal{T}^e$
        \FOR{$e=1$ to $MaxEpisode$}
            \STATE{Initialize the environment and get state $s$}
            \FOR{$t=1$ to $T$}
                \STATE{$a\leftarrow \pi(s;\theta_{actor})$, and execute action $a$}
                \STATE{Obtain reward $r$, next state $s'$, and add this transition to $B$}
                \STATE{Update the critic networks with Eq.\eqref{eq_q}}
                \IF{$t \% 2 == 0$}
                    \STATE{Update $\lambda$ in either linear or adaptive manner}
                    \STATE{Update the actor network with Eq.\eqref{eq8}}
                    \STATE{$\phi_i'\leftarrow \tau \phi_i + (1-\tau)\phi_i'|_{i=1, 2}$}
                    \STATE{$\theta_{actor}'\leftarrow \tau \theta_{actor} + (1-\tau)\theta_{actor}'$}
                \ENDIF
                \STATE{$s\leftarrow s'$}
            \ENDFOR
        \ENDFOR
    \end{algorithmic} 
\end{algorithm}

The proposed pursuit-lock-launch policy learning framework is outlined in Algorithm \ref{alg3}. First, the actor and critic networks are randomly initialized, and the target networks are synchronized. 
In lines 7–8, samples are collected in the simulated environment and added to the replay buffer $B$ of the first-in-first-out store manner.
In lines 9–12, samples are drawn randomly from the replay buffer $B$, and the actor and critic networks are updated.
The loss function for the critic network is as follows:
\begin{equation}
    L_{critic} = E_{(s, a, s', r)\sim D}[(r+min_{i=1, 2}Q_{\phi_i'}(s', \hat{a})-Q_{\phi_i}(s,a))^2|_{i=1, 2}],
    \label{eq_q}
\end{equation}
where $Q_{\phi_i'}$ represents the target Q networks, and $\hat{a}$ denotes the target action at the next state, $\hat{a}\leftarrow\pi(s';\theta_{actor}')$. 
We employ the target networks to stabilize the learning process.
As indicated in Lines 13-14 of Algorithm \ref{alg3}, the target networks are updated in a soft way with weight $\tau$.

\section{Experiments}
We evaluate the proposed learning framework in the constructed simulated combat environment to validate whether the proposed framework can learn successful pursuit-lock-launch policies.
The experiments are designed to answer the following questions: 
\begin{itemize}
    \item How does the proposed learning framework perform compared with state-of-the-art RL and imitation learning approaches? Can the proposed learning framework work against different types of opponent aircraft? (Section 6.1)
    \item Can the proposed framework learn effective pursuit-lock-launch policy? What is the learned policy like? (Section 6.2)
    \item What are the differences between the two modes (linear and adaptive $\lambda$) in the proposed framework? (Section 6.2)
    \item Is the proposed framework robust under different conditions? (Section 6.3)
\end{itemize}

\subsection{Comparative Results}
We compare the proposed learning framework with five state-of-the-art RL and imitation learning methods: 
\begin{itemize}
    \item Behavior cloning (BC), which uses expert data to learn policies in a supervised manner. 
    \item Twin-delayed deep deterministic policy gradient (TD3) method \cite{fujimoto2018addressing}, a RL method designed for continuous action spaces, which improves the deep deterministic policy gradient (DDPG) algorithm \cite{lillicrap2015continuous} with double Q value functions.
    \item Soft Actor-Critic (SAC) algorithm \cite{haarnoja2018soft}, a maximum entropy RL algorithm. 
    \item Expert actor-based SAC approach (E-SAC) \cite{li2023manoeuvre}: The E-SAC algorithm is an imitative RL algorithm. By incorporating high-quality samples generated by experts into the training process of the SAC algorithm, the exploration efficiency and learning speed of the agent are significantly enhanced. Additionally, it can dynamically adjust the proportion of expert samples to exploration samples, ensuring that the learned policy can reach the expert level.
    \item Distributional soft actor-critic (DSAC-v2) \cite{duan2023dsac}: An improved actor-critic approach with distributional value function to enhance the value estimation accuracy.
    
\end{itemize}
Table \ref{table:common_hyperparameters} and Table \ref{table:different_hyperparameters} list the parameter settings used in the experiments. 
Most hyper-parameters are set as the same values for all the methods, as shown in Table \ref{table:common_hyperparameters}, and we tune the replay buffer size for each method to achieve their best performance.
As for the neural network structure, the actor and critic networks, besides the input and output layers, consist of two hidden layers with the Relu activation function. 
The expert datasets for all the baseline methods are collected in AI mode with $20$ successful episodes.
{In AI mode, the agent employs a Proportional-Integral-Derivative (PID) algorithm to regulate its attitude and velocity. The specific implementation involves the following steps: 1) Calculating the altitude deviation and heading angle deviation relative to the target; 2) Setting the autopilot parameters for target altitude and heading angle; 3) Adjusting the angles of the three wings based on the established altitude and heading parameters; 4) Iterating through these steps continuously. Through the aforementioned steps, the agent is capable of maintaining continuous orientation towards the hostile aircraft and initiating engagement upon achieving target lock.}
The expert dataset contains nearly $3e4$ state-action pairs, provided at \url{https://github.com/zrc0622/HIRL4UCAV}.
The episode terminates upon hitting the opponent aircraft down, so the lengths of successful trajectories vary.


\begin{table}
\centering
\caption{Common hyper-parameters among all the methods.}
\label{table:common_hyperparameters}
\begin{tabular}{cc|cc|cc}
\toprule
Hyper-parameters & Value & Hyper-parameters & Value & Hyper-parameters & Value \\
\midrule
Critic lr & 0.001 & Actor lr  & 0.001 & Target update interval & 3         \\ 
Batch size & 128 & Optimizer & Adam & Training iterations of BC & 8e6 \\
$\tau$ & 0.005 & Discount factor & 0.99 & Sizes of hidden layers & $256$, $512$ \\
$T_{\text{Straight}}$ & 1500 & $T_{\text{Serpentine}}$ & 1500 & $T_{\text{Circling}}$ & 1900    \\
\bottomrule
\end{tabular}
\end{table}

\begin{table}
\centering
\caption{Different hyper-parameters for each method.}
\label{table:different_hyperparameters}
\begin{tabular}{c|ccccc}
\toprule
Hyper-parameters & Ours & TD3 & SAC & E-SAC& DSAC-v2 \\
\midrule
$\alpha$    & 10000 & / & / & / & / \\
Size of replay buffer $B$  & 1e5 & 1e5 & 2e5 & 2e5 & 2e5\\
\bottomrule
\end{tabular}
\end{table}

\begin{figure}[htbp]
    \centering
    \includegraphics[width=1\linewidth]{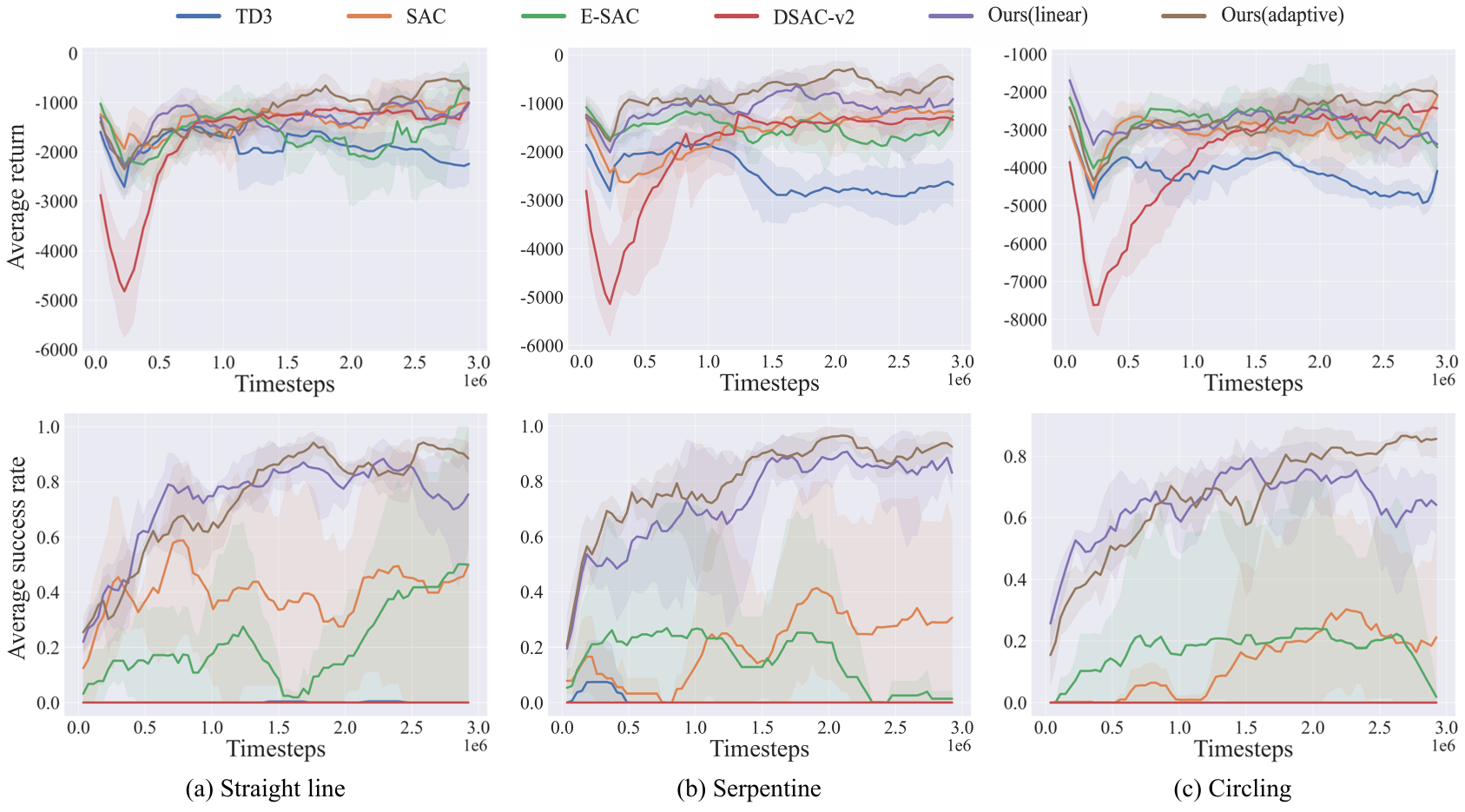}
    \caption{Comparative results of the proposed approach and the baselines against various opponents.}
    \label{exp1}
\end{figure}

{In the experiments, the opponent behaviors have three modes: straight line, serpentine maneuvering, and circling maneuvering.
In the straight mode, the opponent aircraft flies forward at a constant speed. In the serpentine mode, the rudder angle of the opponent repeatedly changes between positive and negative, while the circling maneuvering opponent performs large-angle horizontal turns by adjusting the ailerons and elevators.
The experiment results for these three opponent modes are correspondingly shown in the three columns in Figure \ref{exp1}, where the first row demonstrates the return curves, and the second row demonstrates the success rate curves. For every $25$ training episode, we evaluate the learned policies and calculate the average success rate for $50$ episodes.
Each curve in Figure \ref{exp1} represents the average values of four runs for the same method, with the shaded area indicating the $95\%$ confidence interval. 
The experiments are conducted on the server equipped with Nvidia 2080Ti GPUs, and each run of our approach in Figure \ref{exp1} takes no more than 18 hours.}

The constructed simulator in this study is with high fidelity, since we have designed a realistic action space and task setting. Beyond that, we consider the different properties of various aircraft, and apply real physics models of the aircraft to the simulated engine.
In this realistic simulator, the trained agent has mastered two policies for shooting down the opponent aircraft: (1) directly ramming the opponent aircraft and (2) hitting the opponent aircraft with a missile. Among them, using missiles to shoot down opponent aircraft resulted in higher reward values, which is the correct decision-making we expect the agent to learn. The second row of Figure \ref{exp1} shows the success rate of shooting down opponent aircraft by launching missiles.

As shown in Figure \ref{exp1}, the adaptive weighting method proposed in this study has achieved the highest convergent return and convergent success rate across the three opponent modes.
The comparison of the success rate curves between our method and the TD3 method reveals that the imitation of objectives can improve learning efficiency and success rates\footnote{The success rate curves for TD3 in subfigure (a) and (c) are invisible, since TD3 can hardly learn any successful policy and the success rates are always zero.}. This effect is significant when it is adopted in the form of adaptive weights.
Moreover, compared to other methods, especially the E-SAC method, which also utilizes expert knowledge, our method not only achieves a higher success rate but also exhibits superior stability. This superiority is primarily attributed to the imitated objectives, which directly optimize the policy network and enable rapid improvement of the policy with small samples in a short time, effectively restricting the agent’s exploration space. This enables the agent to explore along the correct path and gradually master key decision-making skills. On the other hand, the E-SAC method indirectly optimizes the policy and Q-value networks by incorporating expert knowledge into the buffer. Although this approach also utilizes expert knowledge, it requires more expert input and results in lower sample efficiency. Our method minimizes the reliance on extensive expert knowledge while ensuring high stability alongside improving success rates by directly optimizing the policy network.
As the imitation learning method, BC, learns in an offline way, so it does not have a learning curve, and therefore we report the convergent performance of BC in Table \ref{table:best_model}.

\begin{table}[]
\caption{Best-performing results averaged over four runs.}
\label{table:best_model}
\resizebox{1.0\linewidth}{!}{
\begin{tabular}{c|cc|cc|cc}
\toprule
\multicolumn{1}{c|}{\multirow{2}{*}{Algorithm}} & \multicolumn{2}{c|}{Straight line opponent} & \multicolumn{2}{c|}{Serpentine opponent} & \multicolumn{2}{c}{Circling opponent} \\
\multicolumn{1}{c|}{}  & \multicolumn{1}{c}{Hit success rate} & \multicolumn{1}{c|}{Return} & \multicolumn{1}{c}{Hit success rate} & \multicolumn{1}{c|}{Return} & \multicolumn{1}{c}{Hit success rate} & \multicolumn{1}{c}{Return} \\ 
\midrule
Ours(adaptive) & $\bf{100.0\%\pm 0.0\%}$ & $\bf{72.3\pm 26.0}$ & $\bf{100.0\%\pm 0.0\%}$ & $\bf{62.5\pm 16.9}$ & $\bf{96.5\%\pm 1.7\%}$ & $\bf{-1187.3\pm 290.3}$ \\
Ours(linear) & $\bf{100.0\%\pm 0.0\%}$ & $36.6\pm 15.8$ & $\bf{100.0\%\pm 0.0\%}$ & $36.4\pm 33.0$ & $96.0\%\pm 2.4\%$ & $-1312.5\pm 316.9$\\
TD3 & $0.0\%\pm 0.0\%$ & $-1033.3\pm 67.3$ & $25.0\%\pm 43.3\%$ & $-1020.9\pm 320.5$ & $0.0\%\pm 0.0\%$ & $-2233.3\pm 211.2$ \\
SAC &$99.0\%\pm 1.7\%$& $30.5\pm 86.3$ & $98.5\%\pm 2.6\%$ &$-270.6\pm 378.1$&$76.0\%\pm 16.3\%$& $-2029.7\pm 1350.1$ \\
E-SAC & $\bf{100.0\%\pm 0.0\%}$ & $-248.9\pm 356.0$ & $72.5\%\pm 42.0\%$& $-601.8\pm 431.8$ &$73.5\%\pm 33.4\%$&$-1598.8\pm 661.1$\\
DSAC-v2 & $0.0\%\pm 0.0\%$& $-791.1\pm 69.6$& $0.0\%\pm 0.0\%$ &  $-825.5\pm 34.3$ & $0.0\%\pm 0.0\%$ & $-1197.8\pm 137.9$ \\
BC & $\bf{100.0\%\pm 0.0\%}$ & $-23.3\pm 15.4$ & $73.0\%\pm 9.2\%$ &$-1102.7\pm 379.9$& $65.0\%\pm 9.1\%$ & $-3055.3\pm 555.9$\\
\hline
Expert & $100.0\%\pm 0.0\%$ & $9.6\pm 16.6$ & $100.0\%\pm 0.0\%$ &$4.3\pm 13.7$& $100.0\%\pm 0.0\%$ &$-931.3\pm 154.5$ \\
\bottomrule
\end{tabular}
}
\end{table}

We select the best-performing results from four runs and calculate their average. The results are presented in Table \ref{table:best_model}. The data in the table is displayed in the format "mean ± standard deviation" for clarity. 
In the aforementioned experiments, we independently implemented all learning algorithms, including the algorithm framework and the GYM environment.
The code of the proposed approach, the learned models, and recorded videos of the learned policies are available at \url{https://github.com/zrc0622/HIRL4UCAV}.

According to Table \ref{table:best_model}, the adaptive weighting method proposed in this study demonstrates superior performance across two key measures (success rates and returns) against three opponent modes. 
Specifically, the TD3 and DSAC-v2 methods show the lowest performance in terms of the hit success rate, which is essentially $0\%$, among all methods. 
In contrast, the method proposed in this study, along with the SAC method, achieve a relatively high hit success rate.
Further comparison between our method and the BC method reveals that, the BC method performs more poorly in terms of both success rate in task completion and reward value compared to our method due to the lack of interaction with the environment and the inability to explore autonomously. 
It is noteworthy that, although the BC method achieves a higher success rate than the TD3 and DSAC-v2 methods in the serpentine maneuvering and circling maneuvering opponent modes, its average returns are significantly lower. This is also attributed to the inability to explore autonomously. When compounding errors occur \cite{xu2020error, ross2010efficient, ross2011reduction}, BC strategy tends to lose control and thus achieves very low returns, whereas TD3 and DSAC-v2 methods can still closely track the opponent by pursuit strategy, thus preventing a rapid decline in returns. 
Comparing the first row and the last row, the proposed approach, benefiting from autonomous exploration, has even accomplished better performance than the expert dataset.

\subsection{Analytical Results}

\begin{figure}[htbp]
    \centering
    \includegraphics[width=0.5\linewidth]{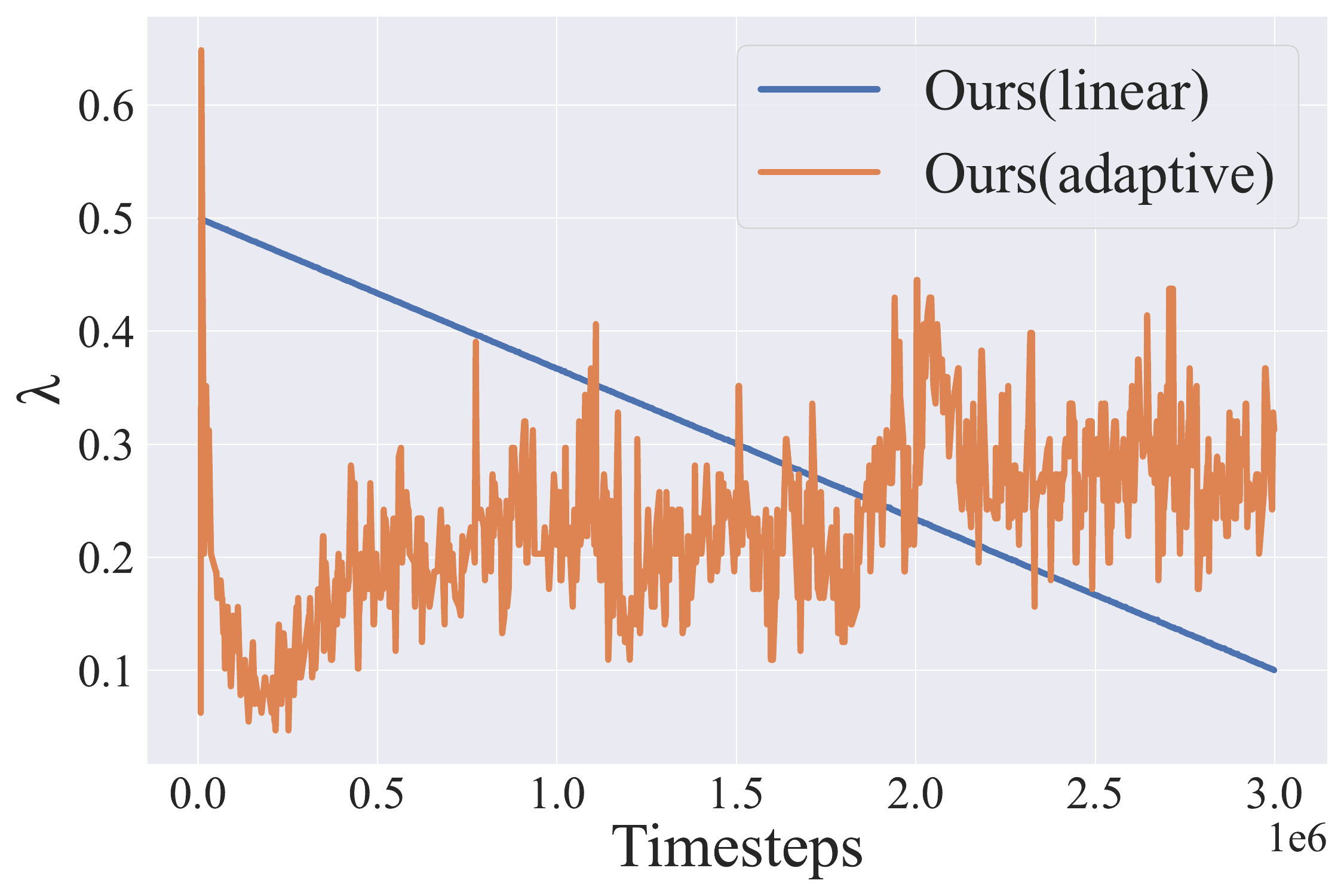}
    \caption{The balancing factor between RL and imitation learning changes with timesteps.}
    \label{fig:lambda}
\end{figure}

We show weight $\lambda$ in the two modes (linear and adaptive) in Figure \ref{fig:lambda}, and find that the value of $\lambda$ is related to the learning performance. As shown in Figure \ref{exp1}, the linear weighting method proposed in this study demonstrates a faster learning rate in the initial stages of training compared to that of the adaptive weighting method. This is primarily attributed to the higher weight assigned to the imitation of objectives by the linear weighting method in the early stages, accelerating the learning of expert knowledge. However, this approach increases the risk of overfitting and limits the model’s generalization ability, resulting in instability and poor convergence performance in training processes. 
Since we show the average $\lambda$ over a batch data in Figure  \ref{fig:lambda}, the $\lambda$ in the adaptive mode is decimal between 0 and 1.

\begin{figure}[htbp]
    \centering
    \includegraphics[width=\linewidth]{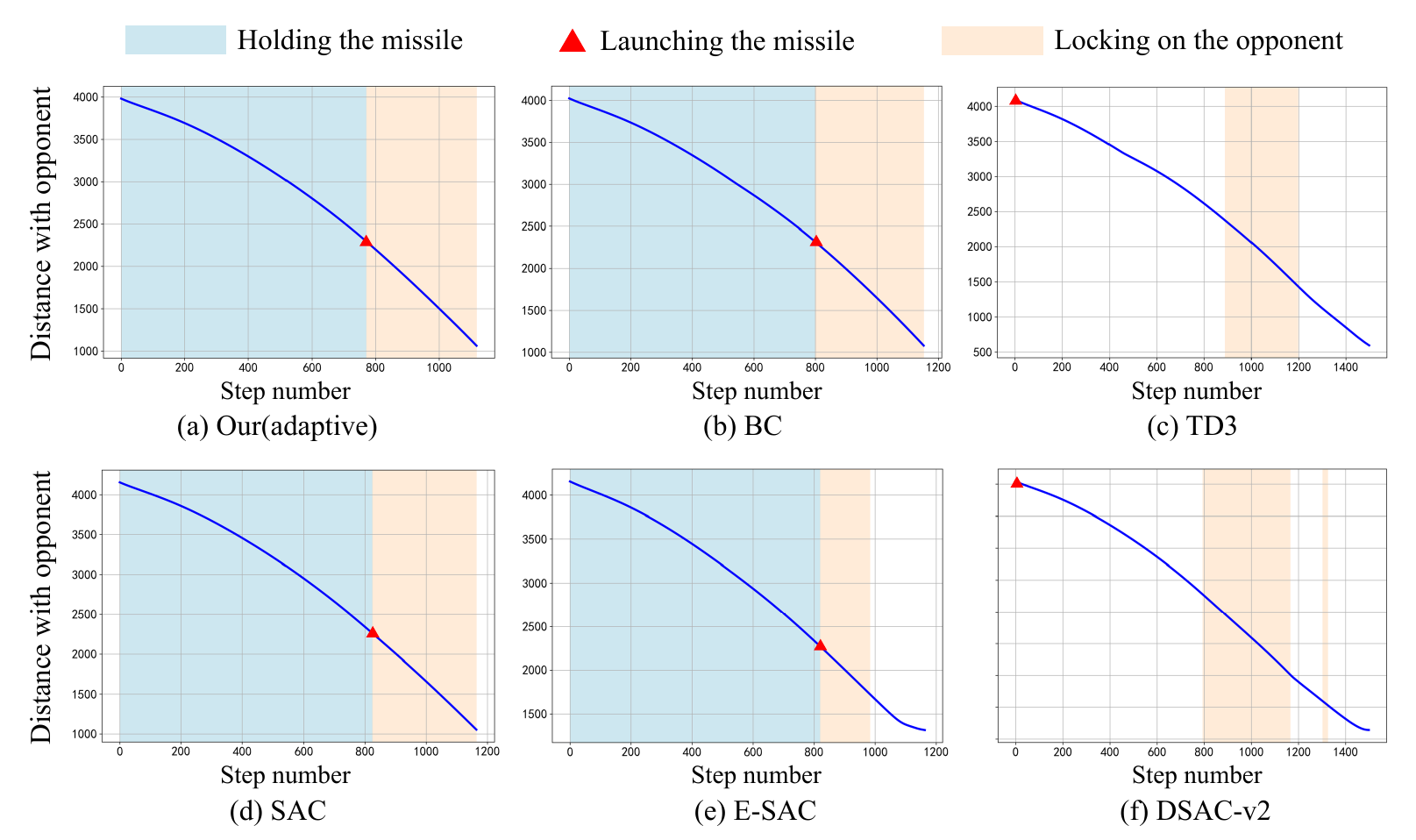}
    \caption{The distance between our aircraft and the opponent in the serpentine opponent mode.}
    \label{fig11}
\end{figure}

Figure \ref{fig11} depicts the distance between our aircraft and the opponent in one episode, and the trajectories are collected by the convergent models learned by the corresponding algorithms in the serpentine opponent mode.
Subfigures in Figure \ref{fig11} annotate the moments of missile launch (red triangles), phases of holding missiles (blue background), and stages of successfully locking onto targets (yellow background). If the first red dot on the curve falls within the intersection of the blue and yellow backgrounds, it indicates that the agent has mastered an effective launch policy. Accordingly, it is observed that the method proposed, along with the SAC, E-SAC, and BC methods, successfully executed missile launches and accurately shot down the opponent aircraft (i.e., mastering the three-dimensional policy of pursuit, lock, and launch). 
Further comparison of the speed of target lock and missile launch among these four methods reveals that the proposed method achieves both with fewer steps, indicating higher attack efficiency. 
The distance curves of the TD3 and DSAC-v2 methods exhibit a downward trend, and there are target locking zones; however, they fail to correctly launch missiles, demonstrating their mastery of only the two-dimensional policy of pursuit and lock. 
Although there is a probability of successful firing in the optimal results of TD3, it does not converge and lacks stability; therefore, we do not consider it to have learned the launch policy. Furthermore, although neither TD3 nor DSAC-v2 can correctly launch missiles, they occasionally manage to succeed by colliding with the opponent during testing, indicating that their pursuit and locking strategies are relatively well-learned.

\begin{figure}[tbp]
\centering
\includegraphics[width=\linewidth]{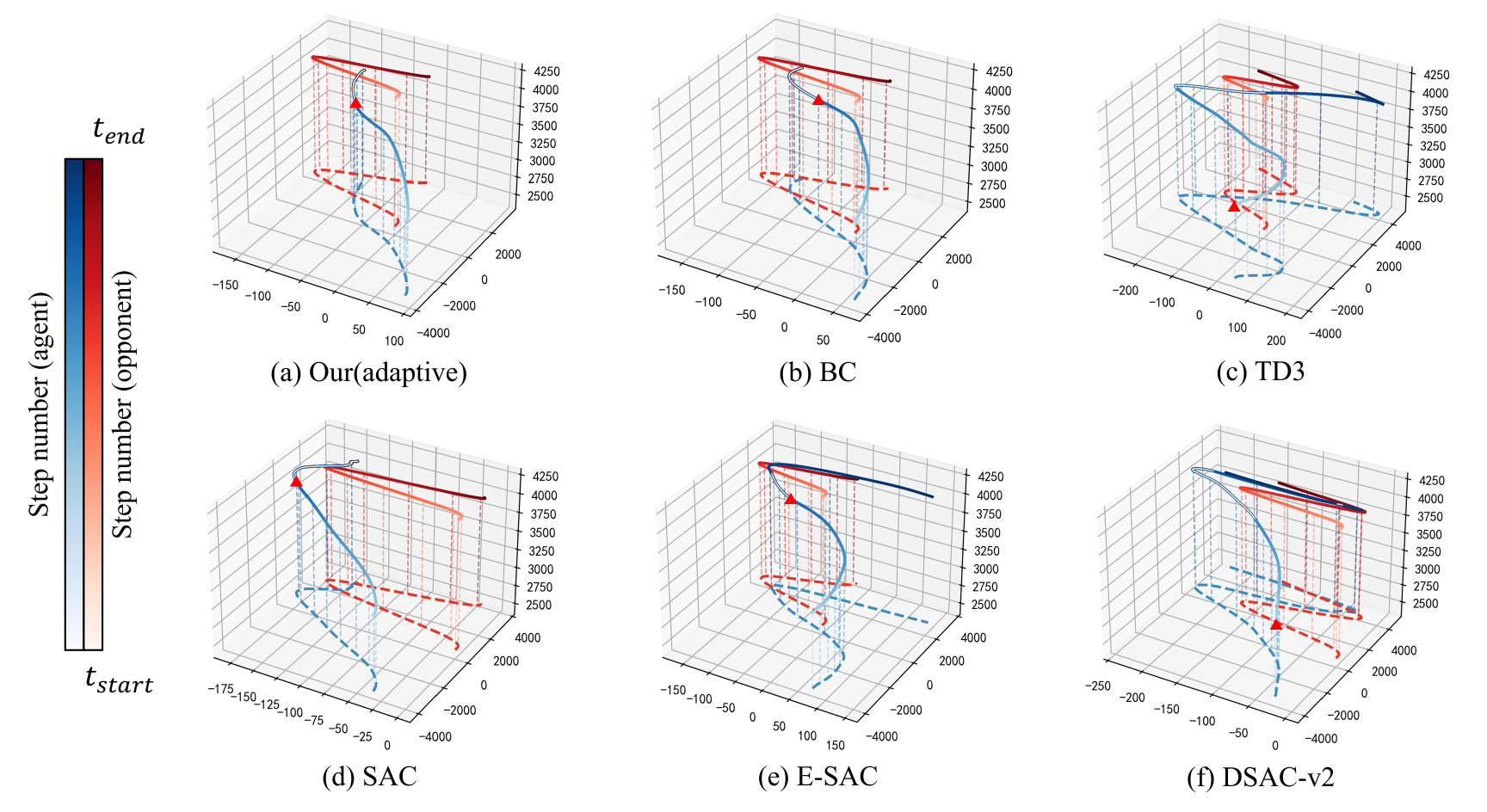}
\caption{The trajectories collected by the learned policy (blue) and the opponent (red) in the serpentine opponent mode.The solid lines represent the 3D trajectories, while the dashed lines are their projections onto the horizontal plane. The lighter color denotes a smaller timestep, and the darker color denotes approaching the end of the episode.The red triangles on the blue trajectories indicate the moments of launch.}
\label{fig12}
\end{figure}

Figure \ref{fig12} corresponds respectively to Figure \ref{fig11}, and Figure \ref{fig12} shows the three-dimensional interaction scenarios between our aircraft and the opponent aircraft using three different methods. The red curves represent the movement trajectories of the opponent aircraft, while the blue curves represent the movement trajectories of our aircraft, with the colors gradually deepening over time. The moments of launch are indicated by red triangles on the blue trajectories, and the yellow areas on the blue trajectories represent the stages of successfully locking onto the opponent. 
Figure \ref{fig12}(a) illustrates the operational effectiveness of the proposed learning framework, i.e., the controlled UCAV swiftly catches up to the opponent aircraft, successfully locking onto it, launching a missile, and accurately hitting the target. 
Note that the blue curve is nearly the shortest path from the start position of our aircraft to the opponent.
In contrast, the UCAV with the TD3 (Figure \ref{fig12}(c)) or DSAC-v2 (Figure \ref{fig12}(f)) policy can pursue the opponent aircraft but fails to shoot it down due to premature missile launches, indicating insufficient launch capabilities. 
Through the above analysis, we systematically summarize the policies learned by all the methods, presented in Table \ref{tab3}.

\begin{figure}[htbp]
\centering
\includegraphics[width=\linewidth]{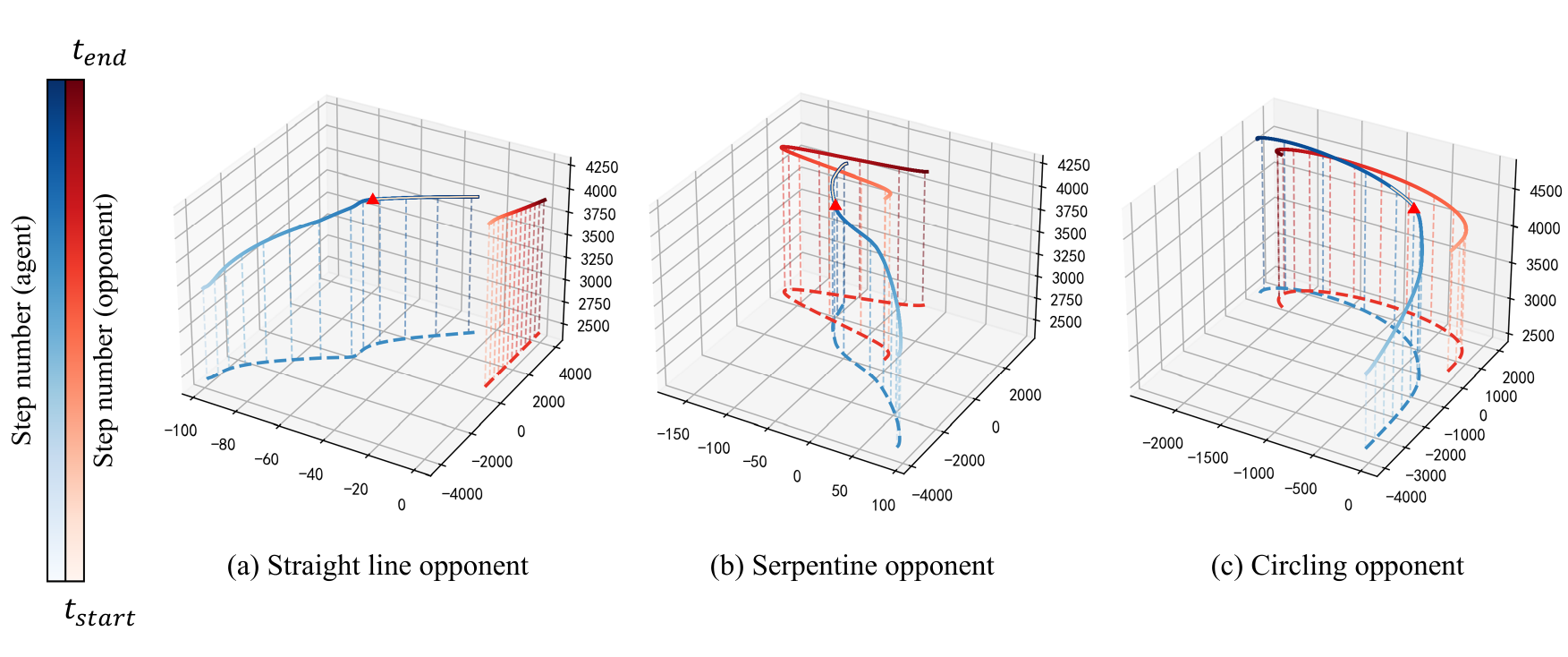}
\caption{The trajectories collected by the policy learned by the proposed method (blue) and the opponent (red) in three opponent modes.The solid lines represent the 3D trajectories, while the dashed lines are their projections onto the horizontal plane. The lighter color denotes a smaller timestep, and the darker color denotes approaching the end of the episode. The red triangles on the blue trajectories indicate the moments of launch.}
\label{fig14}
\end{figure}

In addition to the serpentine mode, we also plotted the results of our method in other opponent modes, as shown in Figure \ref{fig14}. These subfigures demonstrate that the proposed approach can work against various opponent types, shooting down the opponent aircraft in a fast way.  

\begin{table}
\caption{Policy analysis.}
\label{tab3}
\begin{tabular}{cccc}
\toprule
Algorithm& Pursuit & Lock & Launch\\
\midrule
Ours (adaptive)& \color{green}{\checkmark} & \color{green}{\checkmark} & \color{green}{\checkmark} \\
Ours (linear)&  \color{green}{\checkmark} & \color{green}{\checkmark} & \color{green}{\checkmark} \\
TD3&  \color{green}{\checkmark} &  \color{green}{\checkmark} & \color{red}{$\times$} \\
SAC&  \color{green}{\checkmark} &  \color{green}{\checkmark} &  \color{green}{\checkmark}  \\
E-SAC& \color{green}{\checkmark} & \color{green}{\checkmark} & \color{green}{\checkmark} \\
DSAC-v2& \color{green}{\checkmark} & \color{green}{\checkmark} & \color{red}{$\times$} \\
BC& \color{green}{\checkmark} &  \color{green}{\checkmark} & \color{green}{\checkmark} \\

\bottomrule
\end{tabular}
\end{table}

{\subsection{Robustness of the Learned Policy}}

To evaluate the robustness and launch efficiency, we evaluate the models in the unlimited missile supply setting in the serpentine opponent mode. Note that in the training phase, the aircraft only carry one missile.  
Table \ref{tab5} shows the launch efficiency (calculated by the fraction between the hit number and the launch number) of the five methods that have mastered the missile launch policy in the unlimited missile setting, and the methods that cannot effectively launch are omitted.
The proposed learning framework has exhibited outstanding launch efficiency in this experiment, accurately launching missiles regardless of the lock conditions, which demonstrates the effectiveness and reliability of the learned launch policy.

\begin{table}
\caption{Launch efficiency in the unlimited missile setting.}
\label{tab5}
\begin{tabular}{cc}
\toprule
Algorithm& $\frac{\# Hits}{\# Launches}$\\
\midrule
Ours (adaptive)& $\bf{98.8 \%}$\\
Ours (linear)& $98.5 \%$\\
E-SAC & $71.1\%$\\
SAC & $69.0\%$\\
BC & $94.9\%$\\
\bottomrule
\end{tabular}
\end{table}

\section{Conclusion}

This study focuses on the complex multi-dimensional decision-making problems in the UCAV pursuit-lock-launch task.
Leveraging the Harfang3D sandbox, we construct a realistic simulation platform and propose an innovative high-imitative reinforcement learning framework for pursuit-lock-launch policy. This method integrates expert experience with autonomous exploration mechanisms, achieving efficient and precise policy learning. Experimental results demonstrate that the constructed simulation environment serves as an effective platform for conducting reinforcement learning research in aerial combat, enabling UCAVs to handle complex, multistage {WVR engagement} tasks, including target pursuit, target lock, and missile launch in simulated air combat missions. Policies learned with the proposed framework can achieve a hit success rate of up to $100.0\%$. Compared to RL and imitation learning algorithms, the learned policies in this study demonstrate superior success rates and robustness while requiring less expert data.

Our framework can effectively complete multistage tasks, but it exhibits a certain dependence on the quality of expert data, and its training outcomes are influenced by the reward design. In the future, we aim to explore a hierarchical policy to gather representative knowledge from limited expert data, striving to reduce the reliance on expert data quality.

\begin{acks}
This work is supported by the National Natural Science Foundation of China (Grant No.62306088), the Natural Science Foundation of Heilongjiang Province (Grant No.YQ2024F007),  and the Aviation Science Foundation of China (Grant No.2023Z021077001).
\end{acks}

\bibliographystyle{unsrt}
\bibliography{sample-base}










\end{document}